\documentclass[pdflatex,sn-nature]{sn-jnl}

\usepackage{graphicx}%
\usepackage{multirow}%
\usepackage{amsmath,amssymb,amsfonts}%
\usepackage{amsthm}%
\usepackage{mathrsfs}%
\usepackage[title]{appendix}%
\usepackage{xcolor}%
\usepackage{textcomp}%
\usepackage{manyfoot}%
\usepackage{booktabs}%
\usepackage{algorithm}%
\usepackage{algorithmicx}%
\usepackage{algpseudocode}%
\usepackage{listings}%
\usepackage{makecell}%
\usepackage{array}%
\usepackage{tcolorbox}
\usepackage{multirow}
\usepackage{xurl}   
\usepackage{hyperref} 
\usepackage{float}  

\begin{document}

\title[Large Language Models for Large-Scale, Rigorous Qualitative Analysis in Applied Health Services Research]{Large Language Models for Large-Scale, Rigorous Qualitative Analysis in Applied Health Services Research}


\author*[1, 2]{\fnm{Sasha} \sur{Ronaghi}}\email{sronaghi@stanford.edu}
\author[3]{\fnm{Emma-Louise} \sur{Aveling}}\email{eaveling@hsph.harvard.edu}
\author[4]{\fnm{Maria} \sur{Levis}}\email{maria.levis@impactivo.com}
\author[5]{\fnm{Rachel L.} \sur{Ross}}\email{rossrl@stanford.edu}
\author[1,2]{\fnm{Emily} \sur{Alsentzer}}\email{ealsentzer@stanford.edu}
\author*[5,6,7]{\fnm{and Sara} \sur{Singer, for the iPATH investigators}}\email{sara.singer@stanford.edu}

\affil[1]{\orgdiv{Department of Computer Science}, \orgname{Stanford University}, 
  \orgaddress{\street{353 Jane Stanford Way}, \city{Stanford}, \state{CA}, \postcode{94305}, \country{USA}}}
  
\affil[2]{\orgdiv{Department of Biomedical Data Science}, 
  \orgname{Stanford School of Medicine}, 
  \orgaddress{\street{300 Pasteur Dr}, \city{Stanford}, \state{CA}, \postcode{94305}, \country{USA}}}
  
\affil[3]{\orgdiv{Department of Health Policy and Management}, 
  \orgname{Harvard T.H. Chan School of Public Health}, 
  \orgaddress{\street{677 Huntington Ave}, \city{Boston}, \state{MA}, \postcode{02115}, \country{USA}}}

\affil[4]{\orgname{Impactivo LLC},
  \orgaddress{\street{1606 PR-25}, \city{San Juan}, \state{PR}, \postcode{00901}, \country{USA}}}

\affil[5]{\orgdiv{Department of Medicine}, 
  \orgname{Stanford School of Medicine}, 
  \orgaddress{\street{300 Pasteur Dr}, \city{Stanford}, \state{CA}, \postcode{94305}, \country{USA}}}
  
\affil[6]{\orgdiv{Department of Health Policy}, 
  \orgname{Stanford School of Medicine}, 
  \orgaddress{\street{615 Crothers Way}, \city{Stanford}, \state{CA}, \postcode{94305}, \country{USA}}}

\affil[7]{\orgname{Implementing Scalable, PAtient-centered, Team-based, Technology-enabled Care for Adults with Type 2 Diabetes (iPATH) investigators (see Supplementary Information S1)}}


\abstract{
Large language models (LLMs) show promise for improving the efficiency of qualitative analysis in large, multi-site health-services research. Yet methodological guidance for LLM integration into qualitative analysis and evidence of their impact on real-world research methods and outcomes remain limited. We developed a model- and task-agnostic framework for designing human-LLM qualitative analysis methods to support diverse analytic aims. Within a multi-site study of diabetes care at Federally Qualified Health Centers (FQHCs), we leveraged the framework to implement human-LLM methods for (1) qualitative synthesis of researcher-generated summaries to produce comparative feedback reports and (2) deductive coding of 167 interview transcripts to refine a practice-transformation intervention. LLM assistance enabled timely feedback to practitioners and the incorporation of large-scale qualitative data to inform theory and practice changes. This work demonstrates how LLMs can be integrated into applied health-services research to enhance efficiency while preserving rigor, offering guidance for continued innovation with LLMs in qualitative research.
}




\maketitle
\section{Introduction}\label{sec1}
Qualitative approaches are critical to health-services research because they identify the personal and interpersonal experiences, organizational dynamics, and contextual factors that shape how care is delivered and received \cite{Murphy2003}. Yet the demands of large-scale, applied health studies often exceed available time and resources, and findings must be timely to remain relevant for  practitioners \cite{RamanadhanRevetteLeeAveling2021}. Large Language Models (LLMs) may enhance analytic efficiency \cite{AcheampongNyaaba2024, hayes2025conversing}, but their use also risks reducing analysis to superficial, context-blind classifications \cite{Friese2025, Morgan2023} and compromising methodological transparency \cite{AcheampongNyaaba2024, 10.1145/3706598.3713120,ashwin2023usinglargelanguagemodels}.

Methodological guidance for integrating LLMs into real-world qualitative analysis workflows remains limited. Prior work on human-AI collaboration emphasizes maintaining rigor by using AI for specific research tasks and ensuring researchers retain interpretive control \cite{10.1145/3449168, 10.1145/3479856}. However, the flexibility of all-purpose chatbot interfaces like ChatGPT has encouraged more generalized usage across the entire research process \cite{10.1145/3706598.3713120, liao2024llmsresearchtoolslarge, narayanan2024ai}. Such ubiquity risks undermining methodological transparency, a cornerstone of rigor in qualitative analysis \cite{Tracy2010}. Guidance is therefore needed on how to leverage the generality of LLMs and emerging LLM-based analysis tools \cite{Lam_2024, gao} into formalized, task-specific human-LLM qualitative analysis methods, particularly in applied studies where analytic goals, data sources, and most helpful way to report findings are highly context-dependent.

Furthermore, there is a need for greater understanding of how LLM use ultimately influences real-world research methods and outcomes. Novel LLM-based systems for qualitative analysis designed by computer scientists illustrate methodological advancements, but do not focus on integrating LLMs into research studies \cite{Lam_2024, gao}. Qualitative scholars have compared LLMs to human analysts for qualitative coding \cite{Tai2024LLMAidAnalysis, Liu2025QualitativeCodingGPT4, Than2025QualitativeCodingLLM, dunivin2024scalablequalitativecodingllms} and thematic analysis \cite{Wachinger2025PromptsPearlsImperfections, parkington2025human, Naeem2025ThematicAnalysisAI, LevitSaban2025LLMDecisionMaking}. Although they highlight the promise and limitations of LLMs for qualitative analysis, these studies typically prompt LLMs through commercial chat interfaces, limiting their ability to demonstrate LLM usability on large-scale, confidential datasets. Integrating LLMs into ongoing, real-world research studies through extensive collaboration between qualitative and computer science researchers will enable understanding of how LLMs can add value for analyzing large, heterogeneous datasets typical of applied health research. 

In this paper, we propose a framework for designing task-specific human–LLM qualitative analysis methods. We demonstrate how this framework provides methodological guidance for the integration of efficient and rigorous LLM-assisted qualitative analysis within a large, applied health-services research study focused on understanding and improving diabetes care practices at Federally Qualified Health Centers (FQHCs) \cite{iPATH2025}. 

The study spans four research teams from California, Ohio, Massachusetts, and Puerto Rico, involving more than 35 research personnel. Between April and December 2024, researchers conducted a comparative case study across 12 FQHCs, including 167 interviews with clinicians, administrators, patient representatives, and other key stakeholders, to identify organizational conditions and processes that supported or impeded effective diabetes care. The team sought to generate findings for scholarly dissemination, provide actionable feedback to practitioners, and refine a practice transformation intervention designed to improve diabetes care at FQHCs. After refining the intervention, research teams intend to implement the intervention across eight additional FQHC sites and evaluate its impact on patient outcomes. 

Through a collaboration of qualitative and computer science researchers, we utilized LLMs in two distinct qualitative analysis tasks after data collection for the comparative case study across 12 FQHCs. Task~1 involved qualitative synthesis to generate comparative summary reports for feedback to FQHCs, analyzing high-level summaries produced by the research teams about key elements of diabetes care ($\sim${}31{,}200 total words). Task~2 entailed deductive qualitative coding of 167 interview transcripts ($\sim${}8{,}600 total minutes, $\sim$1{,}327{,}000 total words) to refine the planned intervention.

\section{Results}\label{sec2}
\begin{figure}[h]
\centering
\includegraphics[width=1.0\textwidth]{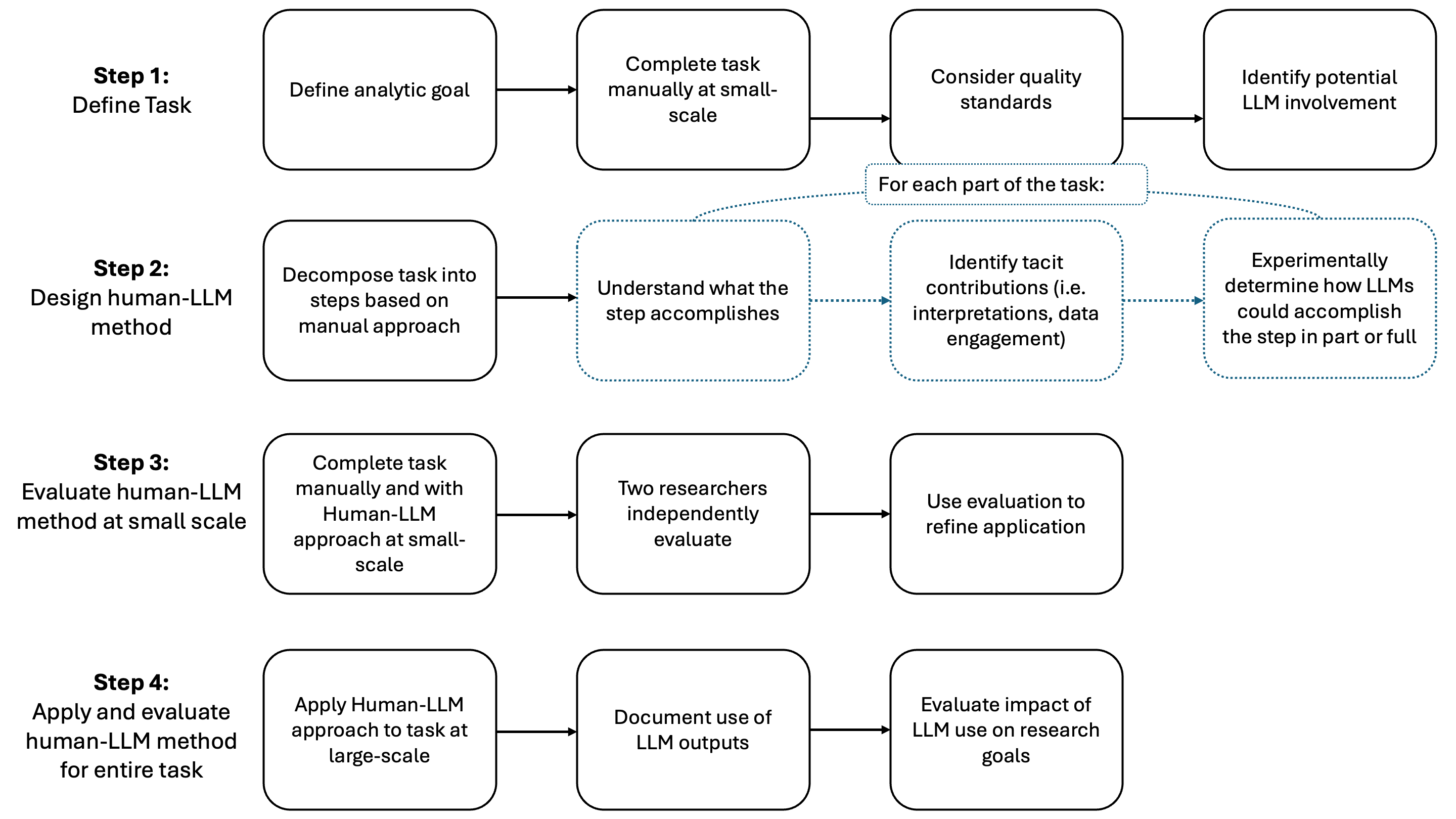}
\caption{The framework we developed and applied for developing task-specific human-LLM qualitative analysis methods.}\label{framework}
\end{figure}

We present a framework for designing task-specific human-LLM qualitative analysis methods  and report results from its application to two analytic tasks: (1) qualitative synthesis to generate comparative summary reports for feedback to FQHCs, and (2) deductive qualitative coding to refine the planned study intervention. For each task, we followed the four steps described below and depicted in Fig.~\ref{framework}.

\bmhead{Step 1: Define Task}
We manually completed the task on a small data sample to clarify goals, finalize output format, document  workflow, and set quality expectations. We also identified which components must be done by qualitative researchers to ensure their familiarity with the data, necessary for future evaluation and interpretation of LLM outputs.

\bmhead{Step 2: Design human-LLM method}
We divided the task into discrete parts to surface actions by qualitative researchers difficult to identify by viewing the task holistically, such as applying context-dependent judgment or drawing on prior knowledge. For each part, we specified its purpose, identified the researcher's tacit contributions, and considered whether to use LLMs, allowing us to test different configurations of human and LLM involvement within each task. 

\bmhead{Step 3: Evaluate human-LLM method at small-scale}

Two researchers compared outputs generated with and without LLM assistance on a small dataset, and their assessments informed team discussion. This approach enabled rigorous evaluation through blending the diverse expertise of the team \cite{levitt2021intersubjective}. 

One challenge in evaluating LLMs for qualitative analysis is the absence of a clear “gold standard,” since replicability is a contested marker of quality in qualitative research \cite{Pownall2024ReplicationQualitativeResponse}. Common quantitative measures such as inter-rater reliability or overlapping thematic coverage \cite{Liu2025QualitativeCodingGPT4, Lam_2024, Wachinger2025PromptsPearlsImperfections} fail to capture whether use of an LLM can achieve the same interpretative depth of insight as qualitative investigators and overlook that appropriate quality markers are contingent on task and study specifics. Instead, we evaluated findings against task-specific goals and established criteria of qualitative rigor such as: grounding in the data, integration of theory and data, alignment with the research question, significance and relevance to the field, and usefulness for practitioners \cite{Tracy2010}. 

These evaluations guided how LLM outputs were integrated and interpreted when applied to the larger dataset. In both tasks, we considered the human-LLM approach sufficient for application to the larger dataset when the variation in outputs between manual and human-LLM analyses resembled the variation we would expect to observe between two human researchers applying the same methods (e.g., showing similar interpretive quality, fit with analytic goals, and utility in the overall analytical process). This benchmark offered a practical way to determine at which steps to incorporate LLMs.

\bmhead{Step 4: Apply and evaluate human-LLM method for entire task}
We documented use of human-LLM output in practice and evaluated its impact on research goals and efficiency.

\subsection{Task 1. Providing Comparative Summary Reports to Participating FQHCs: a Qualitative Synthesis Task}

\bmhead{Step 1: Define Task}
The objective of Task 1 was to transform the qualitative data from the comparative case study into actionable feedback for participating FQHC sites to use for organizational learning and improvement. Specifically, qualitative researchers  sought to generate site-level summaries and cross-site syntheses across 22 care practice domains (e.g. information and communication technology, staff development, patient-provider relationship; see Supplementary Information S2.1), aligned with a widely used management framework \cite{Bodenheimer2002ChronicCarePart2} and the study’s research questions. Qualitative researchers, in consultation with physicians, first developed site-level summaries (3–5 bullet points per domain) without LLM assistance to deepen their familiarity with the data and engage in collaborative reflection about lessons learned. LLMs were then introduced to support cross-site synthesis of each domain. Participating sites each received a report with their own site-level summary and the cross-site synthesis for each domain.

\bmhead{Step 2: Design human-LLM method}
Researchers manually produced cross-site syntheses for each domain first, by grouping site-level summary data into themes, then by synthesizing findings into an actionable summary. This process informed the stages in which we incorporated LLMs. 

Qualitative researchers first grouped each site’s summary bullets for a given domain into themes to facilitate pattern identification across sites, ignoring bullet points deemed less relevant. To mirror this step, we prompted OpenAI’s ChatGPT-4o model to produce output organized into themes. We instructed the LLM to sort original bullet points from all sites within one domain into categorical themes without altering the original data points and creating a “miscellaneous” category for remaining content so all data remained visible to qualitative researchers.

After organizing site summary data into themes, qualitative researchers developed the final cross-site summary for each domain, identifying actionable insights, lessons learned, and creative or good practices. Similarly, we prompted OpenAI’s o1 model to generate a cross-site synthesis based on every site’s data for the domain. We provided the LLM with the task goal, four example manually-derived cross-site syntheses to demonstrate the desired structure and depth—a strategy known as few-shot prompting \cite{brown2020languagemodelsfewshotlearners}—and domain definitions that reflected the management framework and guided how researchers manually interpreted findings. See Supplementary Information S2.2 for prompts.

\bmhead{Step 3: Evaluate human-LLM method at small-scale}
Two qualitative researchers generated cross-site syntheses for two similar domains: information and communication technology and technology for clinical care. Each produced one summary manually and the other with the LLM outputs as reference, ensuring that both approaches were applied to both domains while accounting for individual researcher differences.

The quality of the site summaries developed with and without LLM assistance were comparable in analytic quality and research conclusions, similar to having two qualitative researchers with different perspectives approach the same dataset and question. The AI‑assisted approach reduced time by 30\% for Researcher 1 and 55\% for Researcher 2 (calculated as the difference in time to complete summaries with and without LLM support for the two comparable domains). We attribute this discrepancy to Researcher 1 having greater familiarity with the task and dataset. 

The LLM’s thematically organized output aligned closely with how researchers grouped the data manually. Researchers retained most LLM-generated themes but revised them to be more descriptive, use practitioner-oriented language, and highlight innovation. The miscellaneous category often captured vague or misaligned input data, which, analogous to what inter-rater reliability would reveal, signaled when site teams may have misinterpreted domain definitions. Fig. S3 in Supplementary Information S2.3 illustrates modifications for the information and communication technology domain.

\begin{figure}[h]
\centering
\includegraphics[width=1.0\textwidth]{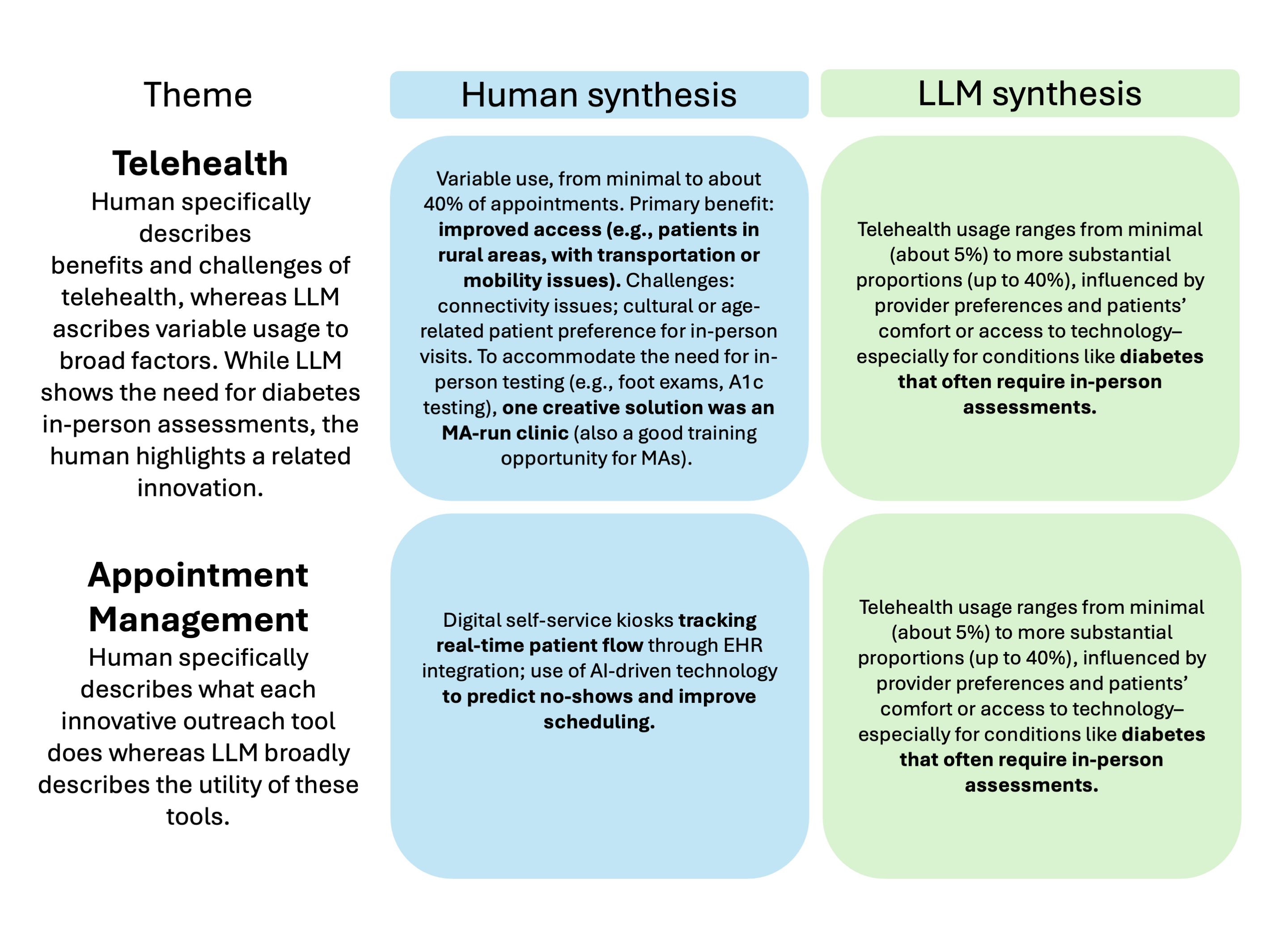}
\caption{Illustrative differences in cross-site synthesis output by human and LLM (independently) for telehealth and appointment management themes within the ``Information and Communication Technology'' domain} \label{fig:ict_differences}
\end{figure}

Comparing the LLM syntheses to those developed manually, we determined that LLMs could not replace researchers’ final interpretation. LLM outputs lacked the nuance and specificity required to be actionable for FQHC sites, shown in Fig. ~\ref{fig:ict_differences}. LLMs also incorporated all input data into the cross-site summary, regardless of whether the data produced novel, useful, and non-redundant insights. In contrast, researchers omitted themes they determined to add little value based on their knowledge of relevant literature and existing practices (example in  Fig. S3 in Supplementary Information S2.3). Additionally, because practices varied at sites, input site summaries did not always align with domain definitions, especially if researchers emphasized different facets of more broadly defined domains. While researchers moved misaligned data to other domains while organizing themes, the LLM included it in the cross-site synthesis (example in  Fig. S3 in Supplementary Information S2.3). These observations highlighted the importance of qualitative researchers making final interpretations of the data to ensure the outputs aligned with the task goal of being actionable and novel for FQHCs. 

\begin{figure}[h]
\centering
\includegraphics[width=1.0\textwidth]{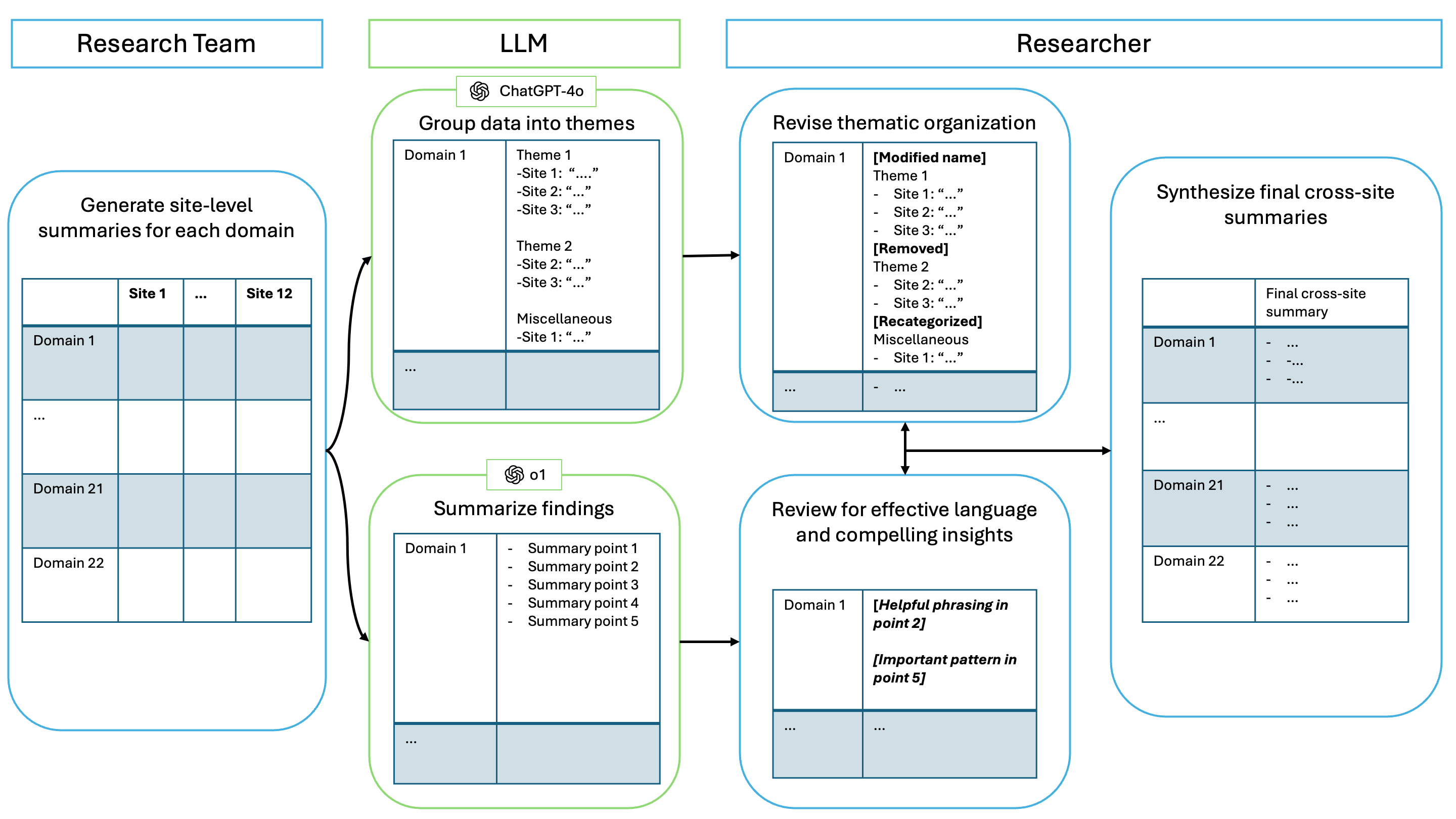}
\caption{\textbf{Final Task 1 human-LLM method.} Step 1, qualitative researchers define domains and create site-level summaries for each one. Step 2, LLM groups data for each domain into themes and provides cross-site synthesis. Step 3, qualitative researcher modifies LLM thematic groups. Step 4, qualitative researchers finalize cross-site synthesis for each domain, highlighting actionable insights and best practices.}\label{task1method}
\end{figure}

The LLM proved useful for organizing site-level data such that qualitative researchers had a comprehensive view of the site summary data, increasing researchers’ confidence that no information was missed in their final interpretation of data. Although LLM-generated summaries were not usable as final output, researchers still found them helpful for identifying patterns across sites and sometimes providing useful language. Based on these findings, we finalized a human-LLM approach (Fig.~\ref{task1method}) where the LLM thematically organizes site summary data for each domain, which the researchers refine and synthesize into a final cross-site synthesis.

\bmhead{Step 4: Apply and evaluate human-LLM method for entire task}
Using LLMs to organize data into patterns lightened cognitive load and enabled researchers to more quickly arrive at a final draft compared to the manual process, ultimately reducing time from data collection to actionable feedback for sites.

\subsection{Task 2. Refining Intervention Design: a Deductive Qualitative Analysis Task}

\bmhead{Step 1: Define Task}
In a prior study \cite{impactivo2019pcmh}, researchers had developed and piloted a practice transformation intervention based on primary care best practices to improve type 2 diabetes patient outcomes at Federally Qualified Health Centers (FQHCs). The objective of Task 2 was to understand the alignment between aspects of diabetes care organization and delivery targeted by the intervention (``practice areas'') and practices observed at the 12 sites in this qualitative comparative case study in order to inform refinement of the intervention for implementation in the next study phase. Specifically, researchers developed a coding framework with 19 broad, semantic, topic summary-type codes covering the practice areas \cite{BraunClarke2006ThematicAnalysis}. LLMs were introduced to generate site-level descriptions for each practice area code supported by evidence from interview transcripts which researchers would analyze to modify components of the intervention. The coding framework is available in Supplementary Information S2.3.

\begin{figure}[h]
\centering
\includegraphics[width=1.0\textwidth]{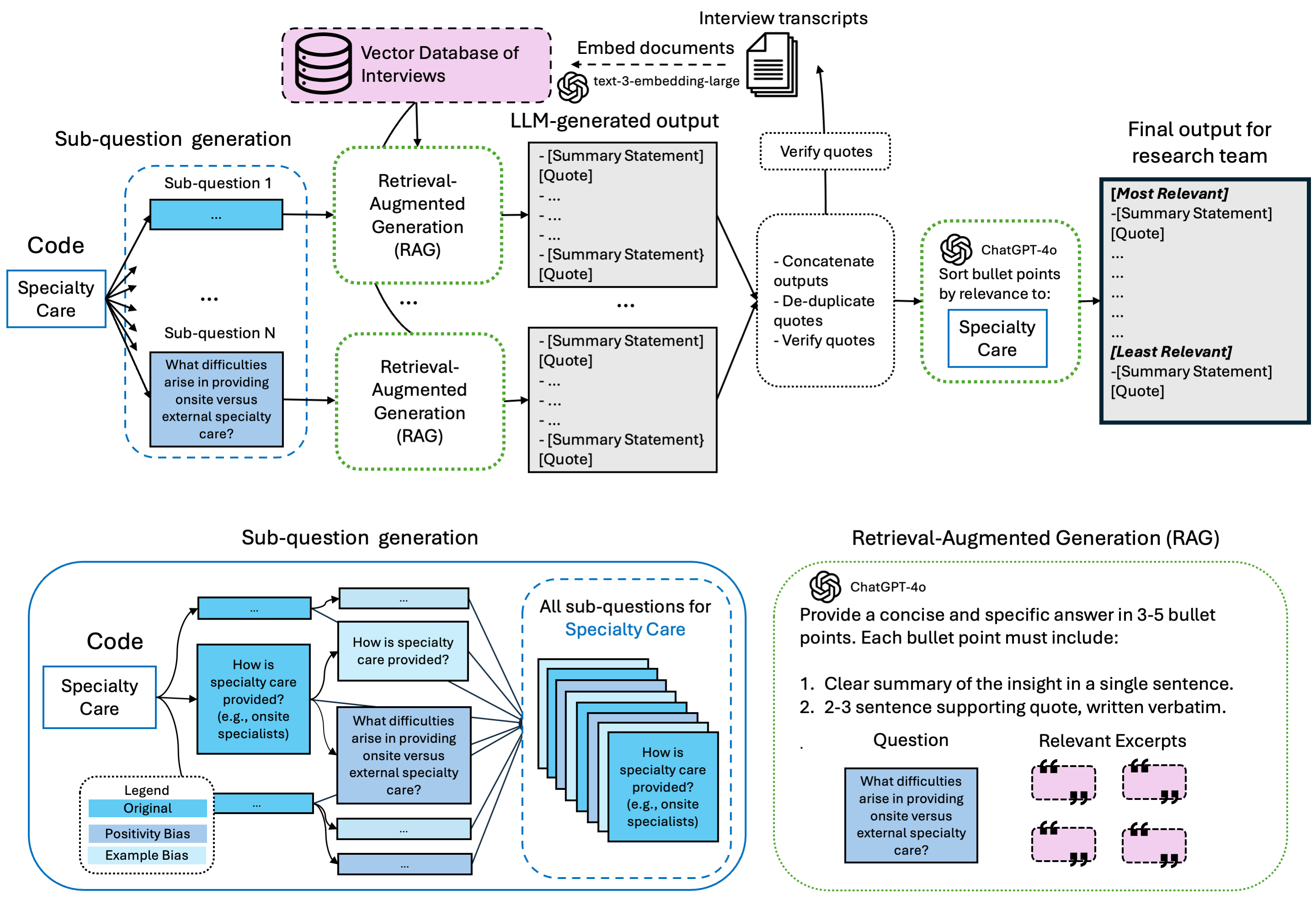}
\caption{\textbf{The human-LLM coding method for each code.} Researchers first generate sub-questions for relevant aspects of the code, then discuss and refine them as a team to ensure alignment. For each sub-question, researchers add two additional questions: 1) if there are examples in the sub-question, the same question without examples (‘example’ bias), and 2) a question on the same topic focused on challenges and barriers (‘positivity’ bias). Next, for each sub-question, we perform Retrieval Augmented Generation: embedding-based retrieval identifies relevant excerpts, and the LLM is prompted to answer using these excerpts. An automated script concatenates LLM outputs for all questions into a single output, and merges duplicate bullet points tied to the same quote, and validates that all quotes appear in the interview text. Finally, the LLM sorts the validated bullet points, which are then provided to the research team.}\label{task2method}
\end{figure}

\bmhead{Step 2: Design human-LLM method}
To inform development of the LLM-integration approach, a researcher coded two interviews from one site for two practice area codes of differing complexity: transportation accessibility (simple) and team-based care (complex). After familiarization with the dataset, the researcher followed a typical deductive coding process for each code: identifying relevant transcript excerpts, then synthesizing and organizing findings based on their judgment of relevance to the code. This process informed iterative development of a human-LLM deductive coding method shown in Fig.~\ref{task2method}. 

Using OpenAI’s ChatGPT-4o, we first aimed to identify transcript excerpts relevant for each code. Prompting the model with an entire interview and code definition yielded vague outputs that missed relevant information, consistent with evidence that LLMs struggle with long-context inputs \cite{liu2023lostmiddlelanguagemodels}. Given our goal to produce site-level analysis, including all interviews per site within a single query was also infeasible: the combined transcripts for each site averaged 157,179 tokens, and 6 of 12 sites exceeded ChatGPT-4o’s 128k token context window limit \cite{openai2024gpt4technicalreport}. We therefore implemented retrieval-augmented generation (RAG) \cite{lewis2021retrievalaugmentedgenerationknowledgeintensivenlp}, which retrieves excerpts before passing them to the LLM. Since abstract concepts such as “team-based care” could not be easily captured through keyword retrieval, we used embedding-based retrieval to surface semantically and contextually similar excerpts without requiring predefined keywords.

The manual approach highlighted that the researcher’s interpretations of code definitions, domain expertise, and dataset familiarity guided how they identified relevant excerpts. For example, the researcher coded excerpts about how teams worked together to provide health education as “team-based care,” even though “health education” was not explicit within the code definition. Connections between seemingly disparate topics are commonly surfaced in qualitative analysis but are not reliably captured by embedding-based retrieval because they depend on tacit knowledge, contextual interpretation, and researcher expertise rather than lexical or semantic similarity. Embedding models surface associations represented in their training data, and, thus, may miss the novel or emergent connections that human analysts often identify. Furthermore, single-vector embeddings, such as OpenAI’s text-embedding-3-large model, cannot capture the full range of relevance relationships between data points, a limitation that amplifies as datasets grow \cite{weller2025theoretical}.

To capture excerpts not explicitly in code definitions but deemed relevant by researchers, three researchers drafted and refined sub-questions for each code through discussion and consensus \cite{richards2018practical}. Each sub-question targeted a single topic to yield specific outputs \cite{openai2024gpt4technicalreport}. Several sub-questions included examples of what researchers expected based on their expertise and familiarization with the dataset, a common practice when developing a qualitative codebook. Unlike humans who may read examples and simultaneously consider novel contexts, we observed that LLMs overfit to examples, likely due to sycophancy bias \cite{perez2022discoveringlanguagemodelbehaviors} from training with reinforcement learning from human feedback \cite{ouyang2022traininglanguagemodelsfollow}. To counter this “example bias,” we posed each question twice—once with and once without examples. We also observed a “positivity bias,” with LLM responses emphasizing positive accounts of what sites were doing and underreporting barriers or challenges. We paired each sub-question with a “counter-perspective” question focused on barriers, similar to the established qualitative practice of seeking “deviant cases” \cite{Anderson2010QualitativeResearch}.

For each sub-question of the code, we performed retrieval-augmented generation (RAG). Reflecting what the team found effective in manual completion of the task and in task 1, we instructed the LLM to produce three to five bullet points, each with a one-sentence summary and illustrative quote. Our experimentation showed that requiring a quote for each statement reduced hallucinations and was more efficient for researchers than reviewing the full set of retrieved excerpts, which were difficult to parse due to abrupt sentence breaks and irrelevant content.

Finally, we aggregated bullet points across sub-questions for each code, removed quotes duplicated across bullet points and verified quotes against transcripts using an automated Python script. We then applied an LLM-as-judge approach to sort bullet points by relevance \cite{gu2025surveyllmasajudge} before providing them to the research team for analysis. LLM prompts are available in Supplementary Information S3.2

\bmhead{Step 3: Evaluate human-LLM method at small-scale}
We compared the LLM-assisted method to manual coding of three interviews each from two FQHC sites (6 total interviews) for four codes of varying levels of conceptual complexity (digital health, patient-provider relationship, defining roles and responsibilities, and patient supports). 

\begin{figure}[h]
\centering
\includegraphics[width=1.0\textwidth]{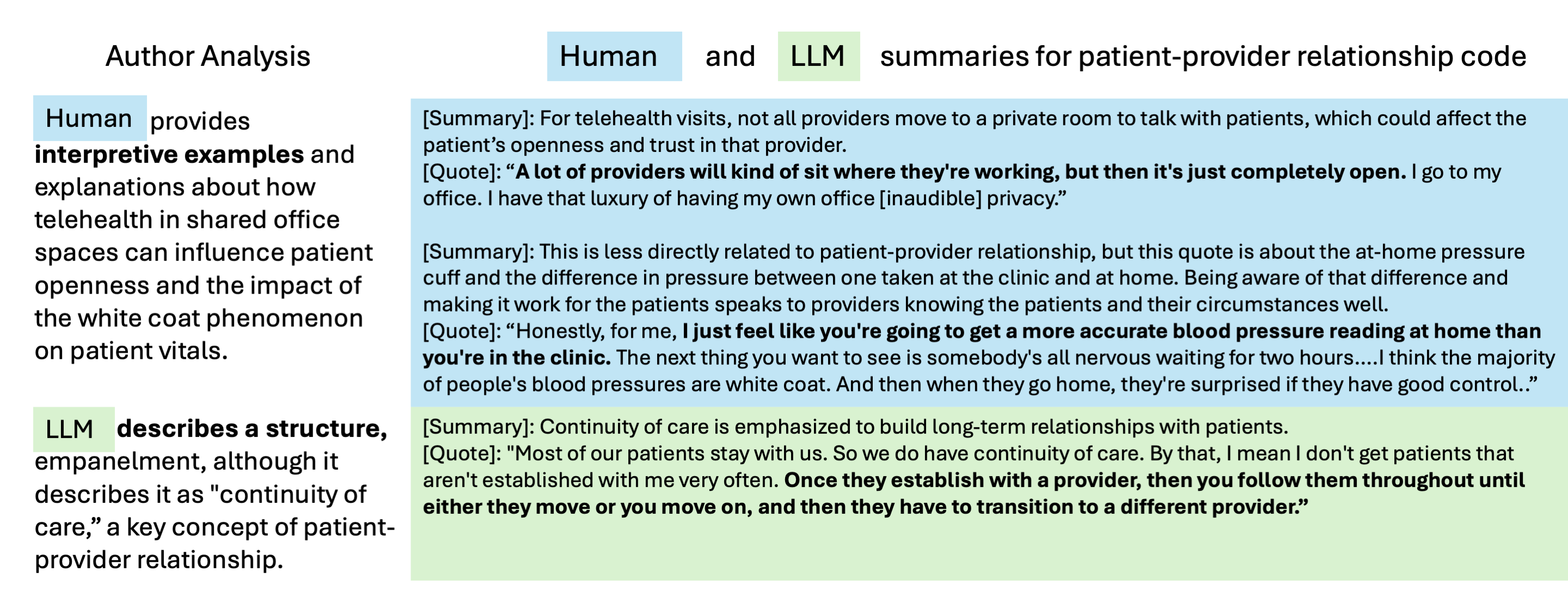}
\caption{\textbf{Different summary statement/quote points brought up by the human and LLMs for the ‘patient–provider relationship’ code}. Human focuses on interpersonal dynamics and LLM on structures and processes. All other summary statement/quote points for this code were very similar.}\label{humanvsllmsummary}
\end{figure}

Overall, the two methods differed in ways similar to how human researchers interpret the same data from different perspectives. Human analysis produced more detailed explanations with fewer quotes, whereas the LLM generated more summary points and supporting quotes, often overlapping thematically due to similarity between sub-questions within each code. Seeing the same theme supported by excerpts from different parts of the dataset increased researchers’ confidence in the LLM-derived findings. Consistent with Levitt \& Saban \cite{LevitSaban2025LLMDecisionMaking}, the human and LLM analyses also identified complementary points: human analysis emphasized interpersonal dynamics, whereas the LLM focused more on structures and processes (Fig.~\ref{humanvsllmsummary}).

\begin{figure}[h]
\centering
\includegraphics[width=1.0\textwidth]{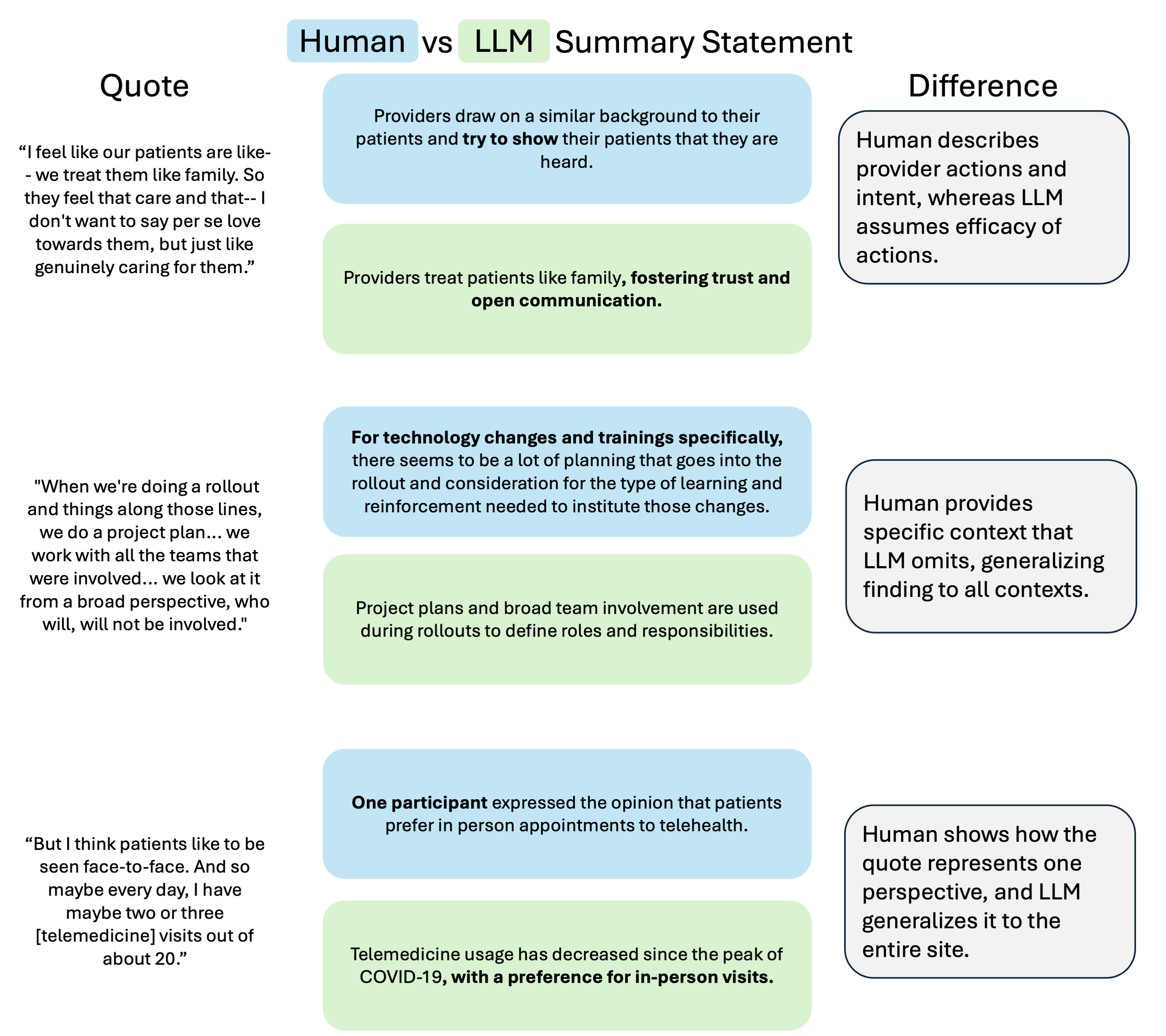}
\caption{\textbf{Same quotes identified by human and LLM, summarized differently} Even though they both selected the same quotes, the LLM often overgeneralized their meaning. This limitation was acute in our dataset, where each interview reflected a distinct perspective requiring nuanced synthesis.}\label{gendiff}
\end{figure}

A key limitation of the LLM output was that our use of RAG restricted its understanding of the dataset to fragmented, decontextualized excerpts, whereas the researcher read through every transcript for the site. Because the team was distributed and site practices varied, not all codes were equally represented: researcher expertise shaped conversation flow, and some codes (e.g., remote work) were irrelevant at certain sites. In these cases, retrieval surfaced low-similarity excerpts, which the LLM treated as relevant without access to the broader dataset, resulting in an output of  tangentially related content. In contrast, the researcher noted reasons for the absence of relevant material. Even when the researcher and LLM both selected the same quotes, the LLM summaries often lacked perspective, overlooked context, and overgeneralized their meaning, reasoning as if excerpts represented the entire site (Fig. \ref{gendiff}). This limitation was acute because, in our dataset,  each interview reflected a distinct perspective requiring nuanced synthesis. These patterns mirror broader limitations of LLMs: trained to produce complete-sounding outputs, they can overstate conclusions even when information is insufficient \cite{openai2024gpt4technicalreport}.

Our evaluation revealed that the human-LLM deductive coding method could sufficiently organize and identify for analysis and interpretation by researchers.  Unlike human-generated coding outputs, however, LLM-generated summary statements require greater researcher judgment to validate and contextualize due to the LLM’s limited contextual understanding of the study goals and dataset. 

\bmhead{Step 4: Apply and evaluate human-LLM method for entire task}

Across the 19 codes, the researchers developed 177 sub-questions, averaging 9 per code (4 – 15 sub-questions per topic). The final LLM output was structured as a matrix with codes as columns and FQHC sites as rows. Each cell contained $\sim$30 bullet points per code per site, with each bullet point pairing a summary statement and a supporting quote. 

Researchers examined the LLM output for each practice-area code to test hypotheses informed by field experience from pilot implementations and interviews of the cross-site case study comparison that they conducted. For example, they observed post-COVID retention challenges at FQHCs. Findings from the “employee wellbeing” and “organizational culture” codes confirmed staff retention as a widespread challenge across all sites. Researchers then incorporated employee wellbeing innovations surfaced through the analysis (e.g. building in opportunities for reflection or checking in) into the intervention, although employee wellbeing was not part of the original intervention or its underlying primary care framework \cite{hahn2014ncqa}. Similarly, researchers hypothesized that fewer sites were practicing empanelment according to its definition (the assignment of patients to providers), despite its centrality to many primary care models, because it was hard to implement in practice. Analysis of the “patient–provider relationship” code confirmed this pattern, prompting a shift of focus from empanelment to strategies maintaining strong patient–provider trust.

Researchers typically reviewed the consolidated outputs for each code across sites and consulted sub-question outputs for specific aspects of interest. Because Step 3 showed that summary statements required validation, they often revisited original transcripts to gain more context, and attribute the output's listing of the source interview for facilitating efficient review. While researchers found the summary statements useful for locating relevant quotes, they did not rely on them heavily for analysis. 

Given a deadline for implementing the refined intervention nine months after collecting data, incorporating findings from all practice areas and sites would have been infeasible without LLMs. In Step 3, the qualitative researcher spent 20 hours coding three transcripts for four codes: 3.5 hours reading through transcripts for entire site, 15 hours coding, and 1.5 hours synthesizing. At that pace, coding one site ($\sim$13 interviews and 19 codes) would require 310 hours (7.75 workweeks) and all 12 sites nearly two years of full-time work (93 workweeks). While not accounting for efficiency gains or individual coding rate variation, these estimates illustrate the substantial time savings of LLMs for qualitative analysis at scale.

\section{Discussion}
We demonstrated how large language models (LLMs) can be integrated into large, multi-site qualitative health-services research through our structured framework that preserves rigor while enabling efficiency. Our framework is especially suitable for applied studies, such as ours, that seek to leverage LLMs to analyze large, heterogeneous datasets and produce outputs tailored to specific goals under time constraints.

Our framework guided our understanding of LLM capabilities within the context of our study, informing our integration of LLMs into the research workflow. We used LLMs primarily for organizational and high-level analytic steps, while reserving theory development and generation of final insights for researchers. In Task 1, LLMs performed on par with researchers in thematically organizing site-level summaries, but they could not judge the relevance or novelty of insights. As a result, researchers relied on LLM-organized data to construct the final cross-site syntheses. In Task 2, LLMs successfully applied a deductive coding framework and identified relevant excerpts across a large dataset, but frequently lacked perspective, overlooked context, or overgeneralized from single data points in summarization. Consequently, researchers treated LLM coding outputs differently from human-generated codes: researchers relied less on interpretations and revisited data more often to ground findings in context. Furthermore, researchers needed access to raw data to demonstrate how their final conclusions were derived, and LLM-generated insights abstracted away both the data and analytic process. This undermined transparency, credibility, reflexivity, and the ability to assess potential biases in LLM-generated findings, such as the reinforcement of assumptions embedded in questions or dominant discourses in training data. Our findings align with prior human–AI collaboration literature underscoring the need for researchers to retain interpretive control \cite{10.1145/3449168, 10.1145/3479856}. We also extend prior work comparing humans and LLM analysis capabilities \cite{Tai2024LLMAidAnalysis, Liu2025QualitativeCodingGPT4, Than2025QualitativeCodingLLM, dunivin2024scalablequalitativecodingllms, Wachinger2025PromptsPearlsImperfections, parkington2025human, Naeem2025ThematicAnalysisAI, LevitSaban2025LLMDecisionMaking} by showing how evaluations of human and LLM abilities informed the design of our human–LLM methods.

While our framework showed that LLMs could perform organizational, high-level analytic steps effectively, these uses still posed risks to the quality of final interpretations. For instance, outputs could become generic, descriptive summaries when data was fragmented or decontextualized. Completing tasks manually on a small scale provided a reference structure that helped us guide LLMs toward outputs clearly traceable to source data, supporting more robust explanatory analysis. Another risk was that researchers could lose familiarity with the data if LLMs replaced activities such as reading transcripts in full. We mitigated this by specifying at the beginning which steps were essential for researchers to remain engaged with data. Furthermore, we observed that variation within our dataset, an inevitable challenge of large, distributed teams, introduced variability in model behavior, making the LLM’s organization less consistent with the researcher’s. Small-scale evaluation helped us identify variability and informed how the team interpreted resulting LLM outputs.

Our interdisciplinary collaboration was essential for operationalizing our framework, and required  iteratively developing, evaluating, and implementing task-specific, custom LLM-based tools and methods, as existing options were not viable. Open-source software often relied on commercial APIs that were not secure for high-risk data \cite{Lam_2024,gao}. Commercial platforms were prohibitively expensive, unable to handle large datasets, and slow to receive institutional approval. Additionally, the proprietary nature of commercial tools reduces transparency in how they implement LLMs, critical for methodological rigor, and limits researchers’ ability to judge appropriate application. Existing options are also rigid: for instance, we could not find a tool that supported specification of output requirements, comparative feedback from matrix inputs, verbatim data reorganization rather than summarization. These limitations matter, especially in applied studies, which often require unique analytic processes and reporting structures, and show how existing tools fail to reflect the well-established principle that task-specific methods are critical for rigorous human–AI collaboration in qualitative analysis  \cite{10.1145/3449168, 10.1145/3479856}.

Iterative collaboration between computer science and qualitative researchers allowed us to overcome these barriers. Computer science researchers enabled our team to implement LLMs in a useful, productive way for our study (e.g., data-processing pipelines based on the study’s existing infrastructure, build reusable prompts and API-based workflows for scalable LLM use, and incorporation of technical innovations such as embedding-based retrieval-augmented generation (RAG)). Qualitative researchers guided technical development to address real analytic bottlenecks and ensured that LLM use did not compromise analytic goals or core quality standards. Additionally, our evaluations were informed both by how qualitative researchers would assess the value of LLM assistance and by technical insights into model behavior and recent advances in computer science. As suggested by prior research \cite{10.1145/3706598.3713120}, having both perspectives was critical to integrating LLMs fruitfully into our ongoing research study, likely explaining why so few projects to date have successfully embedded LLMs into real-world qualitative studies.

Our successful implementation of LLMs through our framework demonstrates their value for analytic efficiency and research outcomes in large-scale, real-world qualitative studies. LLM use accelerated the generation of comparative feedback reports, delivering more timely input to care teams, and enabled the incorporation of insights from 167 transcripts into 19 practice areas of a practice transformation intervention that will be implemented at eight FQHCs in the coming year and may reshape how diabetes care is delivered at FQHCs more broadly. Limitations from our study suggest that advances in LLM long-context reasoning and accessible domain knowledge transfer may further improve LLM capabilities in qualitative analysis, as current models lacked the expertise and capacity to reason across large datasets required for generating insights directly from raw data. Future work should sustain collaboration between qualitative and computer science researchers to ensure methods remain both technically robust and epistemologically grounded. Importantly, our framework can be reapplied across analytic tasks and study contexts, and alongside rapid advances in LLMs, creating opportunities for continued innovations in human–LLM integration for qualitative analysis.

\section{Methods}

We integrated LLMs within the Implementing Scalable, PAtient-centered, Team-based, Technology-enabled Care for Adults with Type 2 Diabetes (iPATH) research study. This multi-year study involves a collaborative network of research teams from Stanford, Harvard, The Ohio State University, and Impactivo, LLC and focuses on practice-relevant research of diabetes care in federally qualified health centers (FQHCs). Investigators collected interview data between April and December 2024 and performed analysis with LLM assistance. The study was approved by Advarra's Institutional Review Board (protocol ID Pro00071432).

To comply with high-risk medical data security standards, we used the Stanford SecureGPT \cite{Ng2024SecureInfrastructureAI} to access commercial LLMs via API, experimenting with models including OpenAI’s ChatGPT-4o and o1, Google Gemini 2.5, Claude Sonnet 3.5, and DeepSeek r1. ChatGPT-4o and o1 consistently followed instructions best and produced the strongest interpretations. 
\bmhead{Task 1}
We developed a Python pipeline to transform an Excel matrix with domains, domain definitions, and site-level summaries into prompts for an LLM and return desired outputs into the study's existing Microsoft Teams environment. Both OpenAI models produced strong thematic organizations, but o1 offered greater depth in summarization. To balance quality, cost, and efficiency, we used ChatGPT-4o for thematic organization and o1 for final cross-site summaries. Supplementary Information S2 contains domains and definitions (S2.1) and  prompts (S2.2).

\bmhead{Task 2}
We built a locally run chat interface to experiment with question types, output formats, and model configurations. Interview documents with metadata (i.e. research team, site, interviewee role, and interviewee role category) were embedded using OpenAI’s text-embedding-3-large model (the largest available through SecureGPT) and stored in a Qdrant vector database \cite{qdrant2025}, enabling pre-search filtering. Users could filter by metadata, adjust model choice (e.g., ChatGPT-4o, o1), parameters (temperature, token limit), retrieval settings (similarity threshold, number of results), and specify output formats before submitting questions. The interface returned both the retrieved excerpts and the LLM outputs, providing transparency into which data informed responses. After finalizing our settings and questions, the interface supported grid-based analysis to compare outputs across metadata partitions, which we used to construct the final output matrix. 

ChatGPT-4o and o1 performed comparably on both the RAG and sorting prompts. ChatGPT-4o was selected for cost and efficiency. To reduce hallucinations, we set model temperature to 0.0. We empirically determined a similarity threshold of 0.4 captured meaningfully relevant excerpts, though we lowered it to 0.3 as some questions returned no results at 0.4. The maximum output length was 4,000 tokens, the limit of Stanford SecureGPT.

Supplementary Information S3 contains each code and its researcher-developed sub-questions (S3.1), prompts (S3.2), and more details on the interface developed to generate analyses (S3.3). 
\backmatter

\section*{Declarations}
\subsection{Author contribution}
S.R. led the study, contributing towards conceptualization of research goals and aims, data analysis, conducting of investigation, development of methodology, project administration, producing software, preparation of presentation of work, and preparation and creation of draft. E.L.A. contributed to research goals and aims, data curation, data analysis, development of methodology, supervision, and revisions of draft. M.L. contributed to development of Task 2 methodology. R.R. and E.A. provided manuscript feedback. S.S. served as principal investigator and contributed towards conceptualization of research goals and aims, data analysis, conducting of investigation, development of methodology, supervision, funding acquisitions, project administration, resources for study completion, and revisions of draft.

\subsection{Funding}
Research reported in this publication was supported by the National Institute On Minority Health And Health Disparities of the National Institutes of Health under Award Number R01MD017870. The content is solely the responsibility of the authors and does not necessarily represent the official views of the National Institutes of Health. The funder played no role in study design, data collection, analysis and interpretation of data, or the writing of this manuscript. 

\subsection{Acknowledgments}

We gratefully acknowledge the organizations that participated in the comparative case study (TBN), and the individuals who participated in study interviews. We thank Cati Brown Johnson and Anna Sophia Lesios of the Evaluative Sciences Unit at Stanford School of Medicine for their assistance with coding interview transcripts as part of our human-LLM methodology testing.

\subsection{Competing interests}
All authors declare no financial or non-financial competing interests. 
\subsection{Data availability}

The datasets generated and/or analyzed during the current study are not publicly available due to the sensitive nature of the interview data and the risk of compromising participant confidentiality. 
De-identified excerpts relevant to the study findings are available from the corresponding author on reasonable request.

\subsection{Code availability}
The underlying code for this study is available in a public Github repository, sronaghi/LLMsinQualAnalysis, and can be accessed via this link \href{https://github.com/sronaghi/LLMsinQualAnalysis}{https://github.com/sronaghi/LLMsinQualAnalysis}.

\subsection{Consent to Participate}
All participants involved in generating analyses for human–LLM comparisons provided informed consent to participate. Participants involved in the interviews provided informed consent under the Implementing Scalable, PAtient-centered, Team-based, Technology-enabled Care for Adults with Type 2 Diabetes (iPATH) study, approved by Advarra’s Institutional Review Board (IRB protocol Pro00071432).

\bibliography{sn-bibliography}

@book{Murphy2003,
  author    = {Murphy, Elizabeth},
  title     = {Qualitative Methods and Health Policy Research},
  edition   = {1st},
  year      = {2003},
  publisher = {Routledge},
  address   = {New York},
  doi       = {10.4324/9781315127873}
}

@article{Tracy2010,
  author  = {Tracy, Sarah J.},
  title   = {Qualitative quality: Eight “big-tent” criteria for excellent qualitative research},
  journal = {Qualitative Inquiry},
  year    = {2010},
  volume  = {16},
  number  = {10},
  pages   = {837--851},
  doi     = {10.1177/1077800410383121}
}

@article{RamanadhanRevetteLeeAveling2021,
  author  = {Ramanadhan, Shoba and Revette, Anna C. and Lee, Rebekka M. and Aveling, Emma L.},
  title   = {Pragmatic approaches to analyzing qualitative data for implementation science: an introduction},
  journal = {Implementation Science Communications},
  year    = {2021},
  volume  = {2},
  pages   = {70},
  doi     = {10.1186/s43058-021-00174-1}
}

@article{AcheampongNyaaba2024,
  author  = {Acheampong, Isaac Owoahene and Nyaaba, Matthew},
  title   = {Review of Qualitative Research in the Era of Generative Artificial Intelligence},
  journal = {SSRN Electronic Journal},
  year    = {2024},
  month   = jan,
  doi     = {10.2139/ssrn.4686920},
  url     = {https://ssrn.com/abstract=4686920}
}

@article{Friese2025,
  author  = {Friese, Susanne},
  title   = {Conversational Analysis with AI — CA to the Power of AI: Rethinking Coding in Qualitative Analysis},
  journal = {SSRN Electronic Journal},
  year    = {2025},
  month   = apr,
  doi     = {10.2139/ssrn.5232579},
  url     = {https://ssrn.com/abstract=5232579}
}

@article{Morgan2023,
  author  = {Morgan, David L.},
  title   = {Exploring the Use of Artificial Intelligence for Qualitative Data Analysis: The Case of ChatGPT},
  journal = {International Journal of Qualitative Methods},
  year    = {2023},
  volume  = {22},
  pages   = {1--10},
  doi     = {10.1177/16094069231211248},
  url     = {https://journals.sagepub.com/doi/full/10.1177/16094069231211248}
}

@article{10.1145/3449168,
author = {Jiang, Jialun Aaron and Wade, Kandrea and Fiesler, Casey and Brubaker, Jed R.},
title = {Supporting Serendipity: Opportunities and Challenges for Human-AI Collaboration in Qualitative Analysis},
year = {2021},
issue_date = {April 2021},
publisher = {Association for Computing Machinery},
address = {New York, NY, USA},
volume = {5},
number = {CSCW1},
url = {https://doi.org/10.1145/3449168},
doi = {10.1145/3449168},
journal = {Proc. ACM Hum.-Comput. Interact.},
month = apr,
articleno = {94},
numpages = {23}
}

@article{10.1145/3479856,
author = {Feuston, Jessica L. and Brubaker, Jed R.},
title = {Putting Tools in Their Place: The Role of Time and Perspective in Human-AI Collaboration for Qualitative Analysis},
year = {2021},
issue_date = {October 2021},
publisher = {Association for Computing Machinery},
address = {New York, NY, USA},
volume = {5},
number = {CSCW2},
url = {https://doi.org/10.1145/3479856},
doi = {10.1145/3479856},
journal = {Proc. ACM Hum.-Comput. Interact.},
month = oct,
articleno = {469},
numpages = {25}
}

@misc{10.1145/3706598.3713120,
      title={Large Language Models in Qualitative Research: Uses, Tensions, and Intentions}, 
      author={Hope Schroeder and Marianne Aubin Le Quéré and Casey Randazzo and David Mimno and Sarita Schoenebeck},
      year={2025},
      eprint={2410.07362},
      archivePrefix={arXiv},
      primaryClass={cs.HC},
      url={https://arxiv.org/abs/2410.07362}, 
}

@misc{liao2024llmsresearchtoolslarge,
      title={LLMs as Research Tools: A Large Scale Survey of Researchers' Usage and Perceptions}, 
      author={Zhehui Liao and Maria Antoniak and Inyoung Cheong and Evie Yu-Yen Cheng and Ai-Heng Lee and Kyle Lo and Joseph Chee Chang and Amy X. Zhang},
      year={2024},
      eprint={2411.05025},
      archivePrefix={arXiv},
      primaryClass={cs.CL},
      url={https://arxiv.org/abs/2411.05025}, 
}

@book{narayanan2024ai,
  title     = {AI Snake Oil: What Artificial Intelligence Can Do, What It Can't, and How to Tell the Difference},
  author    = {Narayanan, Arvind and Kapoor, Sayash},
  year      = {2024},
  publisher = {Princeton University Press},
  address   = {Princeton, NJ}
}

@article{Lam_2024,
  title={Concept Induction: Analyzing Unstructured Text with High-Level Concepts Using LLooM},
  author={Michelle S. Lam and Janice Teoh and James Landay and Jeffrey Heer and Michael S. Bernstein},
  journal={Proceedings of the 2024 CHI Conference on Human Factors in Computing Systems},
  year={2024},
  url={https://api.semanticscholar.org/CorpusID:269214633}
}

@misc{gao,
      title={CollabCoder: A Lower-barrier, Rigorous Workflow for Inductive Collaborative Qualitative Analysis with Large Language Models}, 
      author={Jie Gao and Yuchen Guo and Gionnieve Lim and Tianqin Zhang and Zheng Zhang and Toby Jia-Jun Li and Simon Tangi Perrault},
      year={2024},
      eprint={2304.07366},
      archivePrefix={arXiv},
      primaryClass={cs.HC},
      url={https://arxiv.org/abs/2304.07366}, 
}

@article{Tai2024LLMAidAnalysis,
  author  = {Tai, Robert H. and Bentley, Lillian R. and Xia, Xin and Sitt, Jason M. and Fankhauser, Sarah C. and Chicas-Mosier, Ana M. and Monteith, Barnas G.},
  title   = {An Examination of the Use of Large Language Models to Aid Analysis of Textual Data},
  journal = {International Journal of Qualitative Methods},
  year    = {2024},
  volume  = {23},
  number  = {4},
  pages   = {1--14},
  doi     = {10.1177/16094069241231168},
  url     = {https://doi.org/10.1177/16094069241231168}
}

@article{Liu2025QualitativeCodingGPT4,
  author    = {Liu, Xiner and Zambrano, Andres Felipe and Baker, Ryan S. and Barany, Amanda and Ocumpaugh, Jaclyn and Zhang, Jiayi and Pankiewicz, Maciej and Nasiar, Nidhi and Wei, Zhanlan},
  title     = {Qualitative Coding with GPT-4: Where it Works Better},
  journal   = {Journal of Learning Analytics},
  year      = {2025},
  volume    = {12},
  number    = {1},
  pages     = {169--185},
  doi       = {10.18608/jla.2025.8575},
  url       = {https://learning-analytics.info/index.php/JLA/article/view/8575}
}

@article{Than2025QualitativeCodingLLM,
  author  = {Than, Nga and Fan, Leanne and Law, Tina and Nelson, Laura K. and McCall, Leslie},
  title   = {Qualitative Coding with Generative Large Language Models},
  journal = {Sociological Methods 
            \& Research},
  year    = {2025},
  volume  = {54},
  number  = {3},
  pages   = {849--888},
  doi     = {10.1177/00491241251339188},
  url     = {https://journals.sagepub.com/doi/abs/10.1177/00491241251339188}
}

@misc{dunivin2024scalablequalitativecodingllms,
      title={Scalable Qualitative Coding with LLMs: Chain-of-Thought Reasoning Matches Human Performance in Some Hermeneutic Tasks}, 
      author={Zackary Okun Dunivin},
      year={2024},
      eprint={2401.15170},
      archivePrefix={arXiv},
      primaryClass={cs.CL},
      url={https://arxiv.org/abs/2401.15170}, 
}

@article{Wachinger2025PromptsPearlsImperfections,
  author  = {Wachinger, Jonas and B{\"a}rnighausen, Kate and Sch{\"a}fer, Louis N. and Scott, Kerry and McMahon, Shannon A.},
  title   = {Prompts, Pearls, Imperfections: Comparing ChatGPT and a Human Researcher in Qualitative Data Analysis},
  journal = {Qualitative Health Research},
  year    = {2025},
  volume  = {35},
  number  = {9},
  pages   = {951--966},
  doi     = {10.1177/10497323241244669},
  url     = {https://journals.sagepub.com/doi/10.1177/10497323241244669}
}

@article{parkington2025human,
  title={Human vs. LLM-Based Thematic Analysis for Digital Mental Health Research: Proof-of-Concept Comparative Study},
  author={Parkington, Karisa and Teferra, Bazen G and Rouleau-Tang, Marianne and Perivolaris, Argyrios and Rueda, Alice and Dubrowski, Adam and Kapralos, Bill and Samavi, Reza and Greenshaw, Andrew and Zhang, Yanbo and others},
  journal={arXiv preprint arXiv:2507.08002},
  year={2025}
}

@article{Naeem2025ThematicAnalysisAI,
  author  = {Naeem, Muhammad and Smith, Thomas and Thomas, Laura},
  title   = {Thematic Analysis and Artificial Intelligence: A Step-by-Step Process for Using ChatGPT in Thematic Analysis},
  journal = {International Journal of Qualitative Methods},
  year    = {2025},
  volume  = {24},
  pages   = {1--13},
  doi     = {10.1177/16094069251333886},
  url     = {https://doi.org/10.1177/16094069251333886}
}

@article{LevitSaban2025LLMDecisionMaking,
  author  = {Levit, Naomi Shanwetter and Saban, Mor},
  title   = {When investigator meets large language models: a qualitative analysis of cancer patient decision-making journeys},
  journal = {npj Digital Medicine},
  year    = {2025},
  volume  = {8},
  pages   = {336},
  doi     = {10.1038/s41746-025-01747-3},
  url     = {https://doi.org/10.1038/s41746-025-01747-3}
}

@misc{iPATH2025,
  title        = {R01: Implementing Scalable, Patient-centered Team-based Care for Adults with Type 2 Diabetes and Health Disparities (iPATH)},
  author       = {Sara Singer},
  howpublished = {Rethinking Clinical Trials — demonstration project overview},
  year         = {2025},
  note         = {\url{https://rethinkingclinicaltrials.org/demonstration-projects/ipath/}},
  url          = {https://rethinkingclinicaltrials.org/demonstration-projects/ipath/}
}

@article{Pownall2024ReplicationQualitativeResponse,
  author  = {Pownall, Madeleine},
  title   = {Is replication possible in qualitative research? A response to Makel et al. (2022)},
  journal = {Educational Research and Evaluation},
  year    = {2024},
  volume  = {29},
  number  = {1-2},
  pages   = {1--7},
  doi     = {10.1080/13803611.2024.2314526},
  url     = {https://www.tandfonline.com/doi/full/10.1080/13803611.2024.2314526}
}

@article{Bodenheimer2002ChronicCarePart2,
  author  = {Bodenheimer, Thomas and Wagner, Edward H. and Grumbach, Kevin},
  title   = {Improving Primary Care for Patients With Chronic Illness: The Chronic Care Model, Part 2},
  journal = {JAMA},
  year    = {2002},
  volume  = {288},
  number  = {15},
  pages   = {1909--1914},
  doi     = {10.1001/jama.288.15.1909},
  url     = {https://doi.org/10.1001/jama.288.15.1909}
}

@article{BraunClarke2006ThematicAnalysis,
  author  = {Braun, Virginia and Clarke, Victoria},
  title   = {Using Thematic Analysis in Psychology},
  journal = {Qualitative Research in Psychology},
  year    = {2006},
  volume  = {3},
  number  = {2},
  pages   = {77--101},
  doi     = {10.1191/1478088706qp063oa},
  url     = {https://doi.org/10.1191/1478088706qp063oa}
}

@article{weller2025theoretical,
  title={On the Theoretical Limitations of Embedding-Based Retrieval},
  author={Weller, Orion and Boratko, Michael and Naim, Iftekhar and Lee, Jinhyuk},
  journal={arXiv preprint arXiv:2508.21038},
  year={2025}
}

@misc{perez2022discoveringlanguagemodelbehaviors,
      title={Discovering Language Model Behaviors with Model-Written Evaluations}, 
      author={Ethan Perez and Sam Ringer and Kamilė Lukošiūtė and Karina Nguyen and Edwin Chen and Scott Heiner and Craig Pettit and Catherine Olsson and Sandipan Kundu and Saurav Kadavath and Andy Jones and Anna Chen and Ben Mann and Brian Israel and Bryan Seethor and Cameron McKinnon and Christopher Olah and Da Yan and Daniela Amodei and Dario Amodei and Dawn Drain and Dustin Li and Eli Tran-Johnson and Guro Khundadze and Jackson Kernion and James Landis and Jamie Kerr and Jared Mueller and Jeeyoon Hyun and Joshua Landau and Kamal Ndousse and Landon Goldberg and Liane Lovitt and Martin Lucas and Michael Sellitto and Miranda Zhang and Neerav Kingsland and Nelson Elhage and Nicholas Joseph and Noemí Mercado and Nova DasSarma and Oliver Rausch and Robin Larson and Sam McCandlish and Scott Johnston and Shauna Kravec and Sheer El Showk and Tamera Lanham and Timothy Telleen-Lawton and Tom Brown and Tom Henighan and Tristan Hume and Yuntao Bai and Zac Hatfield-Dodds and Jack Clark and Samuel R. Bowman and Amanda Askell and Roger Grosse and Danny Hernandez and Deep Ganguli and Evan Hubinger and Nicholas Schiefer and Jared Kaplan},
      year={2022},
      eprint={2212.09251},
      archivePrefix={arXiv},
      primaryClass={cs.CL},
      url={https://arxiv.org/abs/2212.09251}, 
}

@misc{ouyang2022traininglanguagemodelsfollow,
      title={Training language models to follow instructions with human feedback}, 
      author={Long Ouyang and Jeff Wu and Xu Jiang and Diogo Almeida and Carroll L. Wainwright and Pamela Mishkin and Chong Zhang and Sandhini Agarwal and Katarina Slama and Alex Ray and John Schulman and Jacob Hilton and Fraser Kelton and Luke Miller and Maddie Simens and Amanda Askell and Peter Welinder and Paul Christiano and Jan Leike and Ryan Lowe},
      year={2022},
      eprint={2203.02155},
      archivePrefix={arXiv},
      primaryClass={cs.CL},
      url={https://arxiv.org/abs/2203.02155}, 
}

@article{Anderson2010QualitativeResearch,
  author  = {Anderson, Claire},
  title   = {Presenting and Evaluating Qualitative Research},
  journal = {American Journal of Pharmaceutical Education},
  year    = {2010},
  volume  = {74},
  number  = {8},
  pages   = {141},
  doi     = {10.5688/aj7408141},
  url     = {https://doi.org/10.5688/aj7408141}
}

@misc{gu2025surveyllmasajudge,
      title={A Survey on LLM-as-a-Judge}, 
      author={Jiawei Gu and Xuhui Jiang and Zhichao Shi and Hexiang Tan and Xuehao Zhai and Chengjin Xu and Wei Li and Yinghan Shen and Shengjie Ma and Honghao Liu and Saizhuo Wang and Kun Zhang and Yuanzhuo Wang and Wen Gao and Lionel Ni and Jian Guo},
      year={2025},
      eprint={2411.15594},
      archivePrefix={arXiv},
      primaryClass={cs.CL},
      url={https://arxiv.org/abs/2411.15594}, 
}

@misc{openai2024gpt4technicalreport,
      title={GPT-4 Technical Report}, 
      author={OpenAI and Josh Achiam and Steven Adler and Sandhini Agarwal and Lama Ahmad and Ilge Akkaya and Florencia Leoni Aleman and Diogo Almeida and Janko Altenschmidt and Sam Altman and Shyamal Anadkat and Red Avila and Igor Babuschkin and Suchir Balaji and Valerie Balcom and Paul Baltescu and Haiming Bao and Mohammad Bavarian and Jeff Belgum and Irwan Bello and Jake Berdine and Gabriel Bernadett-Shapiro and Christopher Berner and Lenny Bogdonoff and Oleg Boiko and Madelaine Boyd and Anna-Luisa Brakman and Greg Brockman and Tim Brooks and Miles Brundage and Kevin Button and Trevor Cai and Rosie Campbell and Andrew Cann and Brittany Carey and Chelsea Carlson and Rory Carmichael and Brooke Chan and Che Chang and Fotis Chantzis and Derek Chen and Sully Chen and Ruby Chen and Jason Chen and Mark Chen and Ben Chess and Chester Cho and Casey Chu and Hyung Won Chung and Dave Cummings and Jeremiah Currier and Yunxing Dai and Cory Decareaux and Thomas Degry and Noah Deutsch and Damien Deville and Arka Dhar and David Dohan and Steve Dowling and Sheila Dunning and Adrien Ecoffet and Atty Eleti and Tyna Eloundou and David Farhi and Liam Fedus and Niko Felix and Simón Posada Fishman and Juston Forte and Isabella Fulford and Leo Gao and Elie Georges and Christian Gibson and Vik Goel and Tarun Gogineni and Gabriel Goh and Rapha Gontijo-Lopes and Jonathan Gordon and Morgan Grafstein and Scott Gray and Ryan Greene and Joshua Gross and Shixiang Shane Gu and Yufei Guo and Chris Hallacy and Jesse Han and Jeff Harris and Yuchen He and Mike Heaton and Johannes Heidecke and Chris Hesse and Alan Hickey and Wade Hickey and Peter Hoeschele and Brandon Houghton and Kenny Hsu and Shengli Hu and Xin Hu and Joost Huizinga and Shantanu Jain and Shawn Jain and Joanne Jang and Angela Jiang and Roger Jiang and Haozhun Jin and Denny Jin and Shino Jomoto and Billie Jonn and Heewoo Jun and Tomer Kaftan and Łukasz Kaiser and Ali Kamali and Ingmar Kanitscheider and Nitish Shirish Keskar and Tabarak Khan and Logan Kilpatrick and Jong Wook Kim and Christina Kim and Yongjik Kim and Jan Hendrik Kirchner and Jamie Kiros and Matt Knight and Daniel Kokotajlo and Łukasz Kondraciuk and Andrew Kondrich and Aris Konstantinidis and Kyle Kosic and Gretchen Krueger and Vishal Kuo and Michael Lampe and Ikai Lan and Teddy Lee and Jan Leike and Jade Leung and Daniel Levy and Chak Ming Li and Rachel Lim and Molly Lin and Stephanie Lin and Mateusz Litwin and Theresa Lopez and Ryan Lowe and Patricia Lue and Anna Makanju and Kim Malfacini and Sam Manning and Todor Markov and Yaniv Markovski and Bianca Martin and Katie Mayer and Andrew Mayne and Bob McGrew and Scott Mayer McKinney and Christine McLeavey and Paul McMillan and Jake McNeil and David Medina and Aalok Mehta and Jacob Menick and Luke Metz and Andrey Mishchenko and Pamela Mishkin and Vinnie Monaco and Evan Morikawa and Daniel Mossing and Tong Mu and Mira Murati and Oleg Murk and David Mély and Ashvin Nair and Reiichiro Nakano and Rajeev Nayak and Arvind Neelakantan and Richard Ngo and Hyeonwoo Noh and Long Ouyang and Cullen O'Keefe and Jakub Pachocki and Alex Paino and Joe Palermo and Ashley Pantuliano and Giambattista Parascandolo and Joel Parish and Emy Parparita and Alex Passos and Mikhail Pavlov and Andrew Peng and Adam Perelman and Filipe de Avila Belbute Peres and Michael Petrov and Henrique Ponde de Oliveira Pinto and Michael and Pokorny and Michelle Pokrass and Vitchyr H. Pong and Tolly Powell and Alethea Power and Boris Power and Elizabeth Proehl and Raul Puri and Alec Radford and Jack Rae and Aditya Ramesh and Cameron Raymond and Francis Real and Kendra Rimbach and Carl Ross and Bob Rotsted and Henri Roussez and Nick Ryder and Mario Saltarelli and Ted Sanders and Shibani Santurkar and Girish Sastry and Heather Schmidt and David Schnurr and John Schulman and Daniel Selsam and Kyla Sheppard and Toki Sherbakov and Jessica Shieh and Sarah Shoker and Pranav Shyam and Szymon Sidor and Eric Sigler and Maddie Simens and Jordan Sitkin and Katarina Slama and Ian Sohl and Benjamin Sokolowsky and Yang Song and Natalie Staudacher and Felipe Petroski Such and Natalie Summers and Ilya Sutskever and Jie Tang and Nikolas Tezak and Madeleine B. Thompson and Phil Tillet and Amin Tootoonchian and Elizabeth Tseng and Preston Tuggle and Nick Turley and Jerry Tworek and Juan Felipe Cerón Uribe and Andrea Vallone and Arun Vijayvergiya and Chelsea Voss and Carroll Wainwright and Justin Jay Wang and Alvin Wang and Ben Wang and Jonathan Ward and Jason Wei and CJ Weinmann and Akila Welihinda and Peter Welinder and Jiayi Weng and Lilian Weng and Matt Wiethoff and Dave Willner and Clemens Winter and Samuel Wolrich and Hannah Wong and Lauren Workman and Sherwin Wu and Jeff Wu and Michael Wu and Kai Xiao and Tao Xu and Sarah Yoo and Kevin Yu and Qiming Yuan and Wojciech Zaremba and Rowan Zellers and Chong Zhang and Marvin Zhang and Shengjia Zhao and Tianhao Zheng and Juntang Zhuang and William Zhuk and Barret Zoph},
      year={2024},
      eprint={2303.08774},
      archivePrefix={arXiv},
      primaryClass={cs.CL},
      url={https://arxiv.org/abs/2303.08774}, 
}

@article{levitt2021intersubjective,
  title={Intersubjective recognition as the methodological enactment of epistemic privilege: A critical basis for consensus and intersubjective confirmation procedures},
  author={Levitt, Heidi M. and Ipekci, Banu and Morrill, Zari and Rizo, Jennifer L.},
  journal={Qualitative Psychology},
  volume={8},
  number={3},
  pages={407--427},
  year={2021},
  publisher={American Psychological Association},
  doi={10.1037/qup0000206}
}

@article{hayes2025conversing,
  author       = {Hayes, Andrew S.},
  title        = {``Conversing'' With Qualitative Data: Enhancing Qualitative Research Through Large Language Models (LLMs)},
  journal      = {International Journal of Qualitative Methods},
  year         = {2025},
  volume       = {24},
  doi          = {10.1177/16094069251322346},
  note         = {Original work published 2025},
  url          = {https://doi.org/10.1177/16094069251322346}
}

@book{impactivo2019pcmh,
  author    = {{Impactivo}},
  title     = {Patient-Centered Medical Home Transformation Series Handbook},
  year      = {2019},
  publisher = {Impactivo}
}

@misc{qdrant2025,
  author       = {Qdrant},
  title        = {qdrant/qdrant},
  year         = {2025},
  howpublished = {\url{https://github.com/qdrant/qdrant}},
  note         = {GitHub repository},
}

@misc{ashwin2023usinglargelanguagemodels,
      title={Using Large Language Models for Qualitative Analysis can Introduce Serious Bias}, 
      author={Julian Ashwin and Aditya Chhabra and Vijayendra Rao},
      year={2023},
      eprint={2309.17147},
      archivePrefix={arXiv},
      primaryClass={cs.CL},
      url={https://arxiv.org/abs/2309.17147}, 
}

@misc{brown2020languagemodelsfewshotlearners,
      title={Language Models are Few-Shot Learners}, 
      author={Tom B. Brown and Benjamin Mann and Nick Ryder and Melanie Subbiah and Jared Kaplan and Prafulla Dhariwal and Arvind Neelakantan and Pranav Shyam and Girish Sastry and Amanda Askell and Sandhini Agarwal and Ariel Herbert-Voss and Gretchen Krueger and Tom Henighan and Rewon Child and Aditya Ramesh and Daniel M. Ziegler and Jeffrey Wu and Clemens Winter and Christopher Hesse and Mark Chen and Eric Sigler and Mateusz Litwin and Scott Gray and Benjamin Chess and Jack Clark and Christopher Berner and Sam McCandlish and Alec Radford and Ilya Sutskever and Dario Amodei},
      year={2020},
      eprint={2005.14165},
      archivePrefix={arXiv},
      primaryClass={cs.CL},
      url={https://arxiv.org/abs/2005.14165}, 
}

@misc{lewis2021retrievalaugmentedgenerationknowledgeintensivenlp,
      title={Retrieval-Augmented Generation for Knowledge-Intensive NLP Tasks}, 
      author={Patrick Lewis and Ethan Perez and Aleksandra Piktus and Fabio Petroni and Vladimir Karpukhin and Naman Goyal and Heinrich K{\"u}ttler and Mike Lewis and Wen-tau Yih and Tim Rocktäschel and Sebastian Riedel and Douwe Kiela},
      year={2021},
      eprint={2005.11401},
      archivePrefix={arXiv},
      primaryClass={cs.CL},
      url={https://arxiv.org/abs/2005.11401}, 
}

@article{Ng2024SecureInfrastructureAI,
  author  = {Ng, Melissa Y. and Helzer, Jennifer and Pfeffer, Marc A. and Seto, Todd and Hernandez-Boussard, Tina},
  title   = {Development of Secure Infrastructure for Advancing Generative AI Research in Healthcare at an Academic Medical Center},
  journal = {Research Square},
  year    = {2024},
  pages   = {rs.3.rs-5095287},
  doi     = {10.21203/rs.3.rs-5095287/v1},
  url     = {https://doi.org/10.21203/rs.3.rs-5095287/v1}
}

@article{hahn2014ncqa,
  author    = {Hahn, K. A. and Gonzalez, M. M. and Etz, R. S. and Crabtree, B. F.},
  title     = {National Committee for Quality Assurance (NCQA) patient-centered medical home (PCMH) recognition is suboptimal even among innovative primary care practices},
  journal   = {Journal of the American Board of Family Medicine},
  year      = {2014},
  volume    = {27},
  number    = {3},
  pages     = {312--313},
  doi       = {10.3122/jabfm.2014.03.130267},
  pmid      = {24808108},
  issn      = {1557-2625},
  month     = {May--Jun}
}

@article{richards2018practical,
  author  = {Richards, K. A. R. and Hemphill, M. A.},
  title   = {A Practical Guide to Collaborative Qualitative Data Analysis},
  journal = {Journal of Teaching in Physical Education},
  volume  = {37},
  number  = {2},
  pages   = {225--231},
  year    = {2018},
  doi     = {10.1123/jtpe.2017-0084},
  url     = {https://doi.org/10.1123/jtpe.2017-0084}
}

@misc{liu2023lostmiddlelanguagemodels,
      title={Lost in the Middle: How Language Models Use Long Contexts}, 
      author={Nelson F. Liu and Kevin Lin and John Hewitt and Ashwin Paranjape and Michele Bevilacqua and Fabio Petroni and Percy Liang},
      year={2023},
      eprint={2307.03172},
      archivePrefix={arXiv},
      primaryClass={cs.CL},
      url={https://arxiv.org/abs/2307.03172}, 
}

\end{document}